\documentclass[journal]{IEEETrans/IEEEtran}

\renewcommand{\cite}{\citep}

\usepackage{amsfonts}
\usepackage{amsmath}
\usepackage{amssymb}
\usepackage{amstext}
\usepackage{bm}

\usepackage{epsf,graphicx}

\usepackage{url}          
\usepackage{cite}         

\title{Efficient Evaluation of the Partition Function of RBMs with
  Annealed Importance Sampling}
\author{Ferran~Mazzanti and Enrique~Romero
\thanks{Ferran~Mazzanti is with the Departament de F\'{\i}sica, Universitat Polit\`ecnica de Catalunya - BarcelonaTech, Spain (email: ferran.mazzanti@upc.edu)}
\thanks{Enrique~Romero is with the Departament de Ci\`encies de la Computaci\'o, Universitat Polit\`ecnica de Catalunya - BarcelonaTech, Spain (email: eromero@cs.upc.edu)}
}

\begin{document}

\maketitle

\begin{abstract}
  Probabilistic models based on Restricted Boltzmann Machines (RBMs)
  imply the evaluation of normalized Boltzmann factors, which in turn require
  from the evaluation of the partition function $Z$. The exact
  evaluation of $Z$, though, becomes a forbiddingly expensive
  task as the system size increases. This even worsens when one
  considers most usual learning algorithms for RBMs, where the exact
  evaluation of the gradient of the log-likelihood of the empirical
  distribution of the data includes the computation of $Z$ at each
  iteration. The Annealed Importance Sampling (AIS) method provides a
  tool to stochastically estimate the partition function of the system.
  So far, the standard use of the AIS algorithm in the Machine
  Learning context has been done using a large number of Monte Carlo
  steps. In this work we show that this may not be required
  if a proper starting probability distribution is employed as the
  initialization of the AIS algorithm. We
  analyze the performance of AIS in both small- and large-sized
  problems, and show that in both cases a good estimation of $Z$
  can be obtained with little computational cost.
\end{abstract}


\maketitle

\section{Introduction}

Restricted Boltzmann Machines
(RBMs)~\cite{smolensky-Restricted-Boltzmann-Machines-1986-PDP}),
constitute a simple variant of the Boltzmann Machines~(BM), where
connections between units in the same layer are forbidden. 
This constrain on the topological architecture of the network allows
performing calculations in a much more efficient way. However, it still
presents the serious drawback of having to
evaluate the normalization constant of the resulting probability
distribution, called the partition function
\begin{equation}
  Z = \sum_{{\bf x},{\bf h}} e^{-E({\bf x},{\bf h})} \ ,
  \label{Z}
\end{equation}
with $E({\bf x}, {\bf h})$ the energy of the different system
configurations. In this expression, the sum runs over all possible
states, which is an exponential function on the number of units. For
this reason, the use of RBMs to build probabilistic models has been
marginal, even being one of the key components of the renaissance of
deep neural network models at the beginning of the 21th
century~\cite{hinton-et-al-DeepBeliefNetworks-2006-NC,
  hinton-salakhutdinov-Auto-Encoders-2006-Science}.

The evaluation of the partition function is a required step in
learning models based on gradient-descent techniques applied to the
log-likelihood of the data. To make things worse, in principle it
must be reevaluated at each iteration of the learning process, because
the weights setting the strength of correlations between units change.
In order to overcome this problem
different approximations have been devised, none of them trying to
evaluate $Z$. The most celebrated choice is the Contrastive
Divergence (CD$_k$) algorithm~\cite{hinton-Contrastive-Divergence-2002-NC},
where all the statistical averages involving $Z$ are replaced by a
single sampling reconstructed after $k$ Gibbs sampling steps. Another
approach uses Parallel Tempering~\cite{EarlDeem_1} to generate
suitable samples, which allow to estimate the required quantities
during the learning
process~\cite{desjardins-et-al-Parallel-Tempering-2010-AISTATS}. However,
this mechanism may require a large amount of intermediate
temperatures, which leads to a large increase in computational
cost.

In fact, the partition function $Z$ is a capital quantity appearing in
many different fields of science. For instance in physics it yields
the Helmholtz free energy, which in turn determines completely the
thermodynamic properties of the system. In this way, it is of
fundamental interest to be able to find good approximations of it.
Among the several possible alternatives~\cite{desjardins-et-al-2011,PhysRevE.91.050101,WangLandau_1},
in this paper we focus on the approach developed by
R.~Neal~\cite{neal-Annealed-Importance-Sampling-1998-TR,
  neal-Annealed-Importance-Sampling-2001-SC}, where an intractable
probability distribution is sampled through a random walk on a space
of intermediate probability distributions. This algorithm turns out to
be efficient in the particular case of the RBM, since the random walk
exploration of the intermediate states can be performed by means of
Gibbs sampling, which is fully
parallelizable~\cite{salakhutdinov-et-al-Annealed-Importance-Sampling-2008-ICML}.
For this reason, the AIS algorithm is the most widely used in the
context of RBMs when an estimate of the partition function is
required~\cite{salakhutdinov-et-al-Annealed-Importance-Sampling-2008-ICML,salakhutdinov-larochelle-2010-AISTATS,
salakhutdinov-hinton-2012-NC,cote-larochelle-InfiniteRBM-2016-NC,xie-et-al-2017,krause-et-al-2018,krause-et-al-2020}. 
However, in most cases the use of AIS
becomes computationally expensive since many large chains of
intermediate probability distributions are used.

In this paper we show that the AIS algorithm can in fact be used to
produce reliable estimates of $Z$ with a small computational cost, even in
realistically large problems, when a proper choice of the starting probability
distribution is suitably selected. This probability distribution can be built from the 
weights of the RBM alone, without the need of a training set. We explore
different approaches to construct this probability distribution, and test them
in different exactly solvable problems. Finally, we compare the results obtained
for realistically large problems with those one gets when using the 
by now standard procedure of Ref.~\cite{salakhutdinov-et-al-Annealed-Importance-Sampling-2008-ICML}.


\section{The partition function of the Restricted Boltzmann Machine
  \label{section-Z_RBM}}

The energy function of a binary RBM with $N_v$ visible units ${\bf x}$ 
and $N_h$ hidden units ${\bf h}$, is defined as:
\begin{equation}
 E({\bf x},{\bf h}) = -{\bf b}^{\rm T}{\bf x} - {\bf c}^{\rm T}{\bf h} -
      {\bf x}^{\rm T}{\bf W}{\bf h} \ .
\label{energy-RBM}
\end{equation}
This expression can be cast as a quadratic form, where visible and
hidden units are organized as row and column vectors preceded by a
constant value of $1$ to account for the bias terms
\begin{equation}
  \tilde {\bf x} = 
  (1 \,x_1\, x_2 \cdots x_{N_v})
  \,\,\,\,\, ,  \,\,\,\,\,
  \tilde {\bf h} =
    (1 \,h_1\, h_2 \cdots h_{N_h}) \ ,
\end{equation}
leading to
\begin{equation}
  E(\tilde{\bf x}, \tilde{\bf h}) = \tilde{\bf x}^{\rm T}
  \left(
  \begin{array}{cc}
    0              & {\bf c} \\
    {\bf b}^{\rm T} & {\bf W} 
  \end{array}
  \right) \tilde{\bf h} 
  \equiv
 \tilde{\bf x}^T \tilde{\bf W} \tilde{\bf h}\ ,
\label{energy-RBM-mat}
\end{equation}
where $\tilde {\bf W}$ is the {\em extended} weights matrix, which 
includes the bias terms.

As usual in energy-based models, the probability of a
state $({\bf x},{\bf h})$ is
\begin{equation}
\label{probability-RBM-xh}
 P({\bf x},{\bf h}) = \frac{e^{-E({\bf x},{\bf h})}}{Z} \ ,
\end{equation}
where the normalization term $Z$ is called the partition function
\begin{equation}
\label{partfun-RBM-1}
 Z = \sum_{{\bf x},{\bf h}} {e^{-E({\bf x},{\bf h})}}.
\end{equation}
The particular form of the energy function (\ref{energy-RBM}) makes
both $P({\bf h}|{\bf x})$ and $P({\bf x}|{\bf h})$ to factorize, and so
it is possible to compute them
  in one step. As a consequence, Gibbs sampling can be computed
efficiently \cite{geman-geman-Gibbs-Sampling-1984-TPAMI}. In addition,
it is also possible to compute efficiently one of the two sums involved in
Eq.~(\ref{partfun-RBM-1}). In this way, for $\{0,1\}$ units, one
has
\begin{equation}
\label{partfun-RBM-2}
Z = \sum_{\bf x} e^{{\bf b}^{T}{\bf x}} \prod_i \left(1 + e^{{\bf c}_i + {\bf
    W}_{\!i}{\bf x}}\right),
\end{equation}
where index $i$ runs over the whole set of hidden units, and ${\bf
  W}_{\!i}$ is the $i$th row of ${\bf W}$. However, the evaluation of
$Z$ is still computationally prohibitive when the number of input and
hidden variables is large, since it involves an exponentially large
number of terms. 
For that reason, RBMs are computationally hard to evaluate or 
simulate accurately~\cite{long-serveido-RBMs-Hard-Evaluate-Simulate-2010-ICML}.


\section{Annealed Importance Sampling
  \label{section-AIS}}

Annealed Importance Sampling was developed by R.~Neal in the late
90's~\cite{neal-Annealed-Importance-Sampling-1998-TR,
  neal-Annealed-Importance-Sampling-2001-SC}. AIS allows sampling from
a probability distribution that would be otherwise intractable. Assume
we want to evaluate the average value of some quantity $\alpha({\bf x})$
over a probability distribution $p({\bf x})$, This computation can be
very inefficient due to two main reasons. On one hand, the
probability distribution $p({\bf x})$ can be impossible to sample
because the exact form of $p({\bf x})$ is not known, as it happens,
for instance, in many quantum physics problems~\cite{ceperley-1995,kosztin-et-al-1996,sarsa-et-al-2000}.
On the other hand, the number of samples required to obtain an
accurate estimate of the average value of $\alpha({\bf x})$ may be
unreasonably large. In order to deal with these problems, one usually
resorts to some form of Importance Sampling, where the exploration of
the space is guided by a known and suitable probability distribution
$q({\bf x})$ ~\cite{srinivasan-2002}. Without loss of
generality, the average value of $\alpha({\bf x})$ over $p({\bf x})$ can
be written in the form

\begin{equation}
  \langle \alpha \rangle = \int d{\bf x}\,p({\bf x}) \alpha({\bf x}) \ .
\label{xixi}
\end{equation}
This quantity can be approximated by sampling from $p({\bf
  x})$. However, in many situations, as mentioned above, it is not
possible to sample efficiently or directly from $p({\bf x})$. 
This same quantity can be evaluated using an Importance Sampling
distribution $q({\bf x})$ as
\begin{equation}
  \langle \alpha \rangle = \int d{\bf x}\,q({\bf x})
  \left(
  { p({\bf x}) \alpha({\bf x}) \over q({\bf x}) } 
  \right) \ .
  \label{xi-is}
\end{equation}
In this case, samples are drawn from the probability distribution $q({\bf
  x})$, and
$p({\bf x}) \alpha({\bf x})/q({\bf x})$ is accumulated to estimate $\langle \alpha\rangle$.
Importance Sampling is employed to reduce the variance
of the estimator, or to reduce the number of samplings needed to
achieve the same statistical accuracy. 

In any case, Importance Sampling can only be performed when a suitable
$q({\bf x})$ is at hand, but that may not always be the case. The AIS
method allows building a suitable $q({\bf x})$ starting from a trivial
probability distribution, and annealing with a set of intermediate
distribution corresponding to decreasing temperatures.

As explained in~\cite{neal-Annealed-Importance-Sampling-1998-TR,
  neal-Annealed-Importance-Sampling-2001-SC}, in order to
estimate $\langle\alpha\rangle$ 
starting from a trivial $p_0({\bf x})$, 
one builds a chain of intermediate distributions $p_i({\bf x})$ that
interpolate between $p_0({\bf x})$ and 
$p_n({\bf  x})=p({\bf x})$. A common scheme to define the intermediate
distributions is to set
\begin{equation}
  p_j({\bf x}) = p_0({\bf x})^{1-\beta_j}
  p_n({\bf x})^{\beta_j}
  \ ,
\label{p_AIS}
\end{equation}
with $0 = \beta_0 < \beta_1 < \cdots < \beta_n = 1$. The approach used
in AIS is to turn the estimation of $\langle \alpha\rangle$ into a
multidimensional integration of the form


\begin{equation}
  \langle\alpha\rangle = \int d{\bf x}_1 \cdots d{\bf x}_n \,
  g({\bf x}_1, \cdots, {\bf x}_n)
                    { f({\bf x}_1, \cdots, {\bf x}_n) \over
                      g({\bf x}_1, \cdots, {\bf x}_n) }
                    \alpha({\bf x}_n)
                    \ ,
  \label{xi_AIS}
\end{equation}
where
\begin{eqnarray}
  f({\bf x}_1, \cdots, {\bf x}_n) \!\!\! & = & \!\!\! p_n({\bf x}_n)
  \prod_{j=1}^{n-1}
  \tilde T_j({\bf x}_{j+1}, {\bf x}_j)
  \label{fn} \\
  g({\bf x}_1, \cdots, {\bf x}_n) \!\!\! & = & \!\!\! p_0({\bf x}_1)
  \prod_{j=1}^{n-1}
  T_j({\bf x}_j, {\bf x}_{j+1})  
\end{eqnarray}
are normalized joint probability distributions for the set of
variables $\{ {\bf x}_1, \ldots, {\bf x}_n\}$.
In these expressions $T_k({\bf x},{\bf y})$ represents a transition
probability of moving from state ${\bf x}$ to state ${\bf y}$, which
asymptotically leads to the equilibrium probability $p_k({\bf z})$.
In the same way, $\tilde T_k({\bf y},{\bf x})$ represents the reversal
of $T_k({\bf x},{\bf y})$.  The detailed balance condition implies
that the transition probabilities fulfill the relation
\begin{equation}
  \tilde T_j({\bf y}, {\bf x}) = T_j({\bf x}, {\bf y})
         {p_j({\bf x}) \over p_j({\bf y})} 
  \label{det_balance}
\end{equation}  
in order to be able to sample the space ergodically~\cite{Amit_89}.
Therefore, $\langle\alpha\rangle$ can be estimated from Eq.~(\ref{xi_AIS}) because:
a) the ratio appearing in Eq.~(\ref{xi_AIS}), 
which represents the importance sampling weights, becomes the product of the
ratios of the intermediate probability distributions
\begin{equation}
  { f({\bf x}_1, \ldots, {\bf x}_n)
    \over
    g({\bf x}_1, \ldots, {\bf x}_n)
  } =
  \prod_{k=1}^n { p_k({\bf x}_k) \over p_{k-1}({\bf x}_k) } \ ,
  \label{fg_1}
\end{equation}
and b) $g({\bf x}_1, \ldots, {\bf x}_n)$ is easily sampled from the trivial $p_0({\bf x})$.

In practice, one uses $g({\bf x}_1, \ldots, {\bf x}_n)$ to generate
$N_s$ samples of all the intermediate distributions, such that for
every set of values $\{ {\bf x}_1^i, {\bf x}_2^i, \ldots, {\bf x}_n^i
\}$ (with $i$ spanning the range $[1,N_s]$) 
one gets a set of weights $\{\omega_i\}$ upon substitution in
Eq.~(\ref{fg_1}). In this way, $\langle \alpha\rangle$ is estimated
according to
\begin{equation}
  \langle \alpha\rangle \sim
          { \sum_{i=1}^{N_s} \omega_i \alpha({\bf x}_n^i) \over
            \sum_{i=1}^{N_s} \omega_i } \ ,
  \label{xi_AIS_b}
\end{equation}
with
\begin{equation}
\omega_i = \prod_{k=1}^n { p_k({\bf x}^i_k) \over p_{k-1}({\bf x}^i_k) } \ .
\label{omegai}
\end{equation}
This expression can be written in term of the unnormalized
probabilities $\tilde p_k({\bf x})=Z_k p_k({\bf x})$ as
\begin{equation}
  \omega_i = {Z_0 \over Z_n}
    \prod_{k=1}^n { \tilde p_k({\bf x}^i_k) \over \tilde p_{k-1}({\bf
        x}^i_k) } = {Z_0 \over Z_n} \tilde\omega_i \ ,
\label{omegai_k}
\end{equation}
which defines the set of importance weights $\{\tilde\omega_i\}$
obtained from the product of the ratios of the unnormalized
probabilities. Notice that $\tilde \omega_i$ is an accessible quantity,
while $\omega_i$ is not, just because one does not have access to $Z_n$.
One important consequence of this formalism is that a simple estimator
of the partition function $Z_n$ associated to the distribution
$p_n({\bf x})=p({\bf x})$ is directly given by the average value
\begin{equation}
  {Z_n \over Z_0} \sim {1\over N_s} \sum_i \tilde\omega_i \ .
  \label{Z_AIS_wi}
\end{equation}

Usually, though, the values of $\tilde\omega_i$ are so large that one
typically has to draw samples of $\log(\tilde\omega_i)$ instead.
%
In this way, one defines a set of
$Z_0$-normalized AIS samples $s_i = \log(\tilde\omega_i)+\log(Z_0)$, such that
\begin{equation}
\log(Z_{\rm AIS}) = 
\log\left\langle Z_n \right\rangle_{s} = 
\log\left[{1\over N_s} \sum_i e^{s_i}\right]
\approx \log(Z_n) \ ,
\label{log_mean_exp}
\end{equation}
which is the logarithmic mean of the exponentiated samples. Notice that this
value is different from the mean of the samples $s_i$, although in many
cases is similar. In fact, these two quantities tend to be the same when the
variance of the set of samples is small compared to the mean value. In other situations,
the nonlinear character of the operation in Eq.~(\ref{log_mean_exp}) makes the result be 
dominated by the largest samples, to the point that, in the extreme case, the largest sample
exhausts the total sum.

\section{Efficient use of AIS in RBMs
  \label{section-AIS_RBM}}

For the special case of the RBM, the equilibrium Boltzmann
distribution associated to the visible layer is given by
\begin{equation}
  p({\bf x}) =   \sum_{\bf h} p({\bf x}, {\bf h}) =
  {1\over Z}\sum_{\bf h} e^{-E({\bf x},{\bf h})}
  \label{Boltz_RBM}
\end{equation}
with the energy function of Eq.~(\ref{energy-RBM}).  In the spirit of
AIS, the partition function associated with $p({\bf x})$ can be
obtained from a chain of intermediate probability distributions.
An easy-to-sample distribution $p_0({\bf x})$, built from a RBM model
containing only visible bias terms ${\bf B}$, turns out to be a
convenient starting point.
In this way one generates a family of energy functions
\begin{equation}
  -E_k({\bf x},{\bf h}) = (1-\beta_k) {\bf B}^T {\bf x}
  +\beta_k \left( {\bf b}^T{\bf x} + {\bf c}^T{\bf h} +
      {\bf h}^T{\bf W}{\bf x}\right) \ ,
\label{energy-RBM_AIS}
\end{equation}
with $\beta_k=k/n$ and $k=0, 1, \ldots, n$. Notice that this
prescription is not exactly the same as the one reported in Eq.~({\ref{p_AIS}}), although
this is not relevant since the AIS algorithm does not impose
a specific scheme. As expected, one recovers $p({\bf x})$ for $k=n$. 

In this scheme, 
$E_0({\bf x},{\bf h})={\bf B}^T {\bf x}$, which makes $p_0({\bf x})$ 
a probability distribution that is trivial to sample, with 
$Z_0=2^{N_h}\prod_{j=1}^{N_v}\left(1+e^{B_j}\right)$. 
In this sense, this procedure is similar 
to the one presented in~\cite{salakhutdinov-et-al-Annealed-Importance-Sampling-2008-ICML}.
According to Eq.~(\ref{Z_AIS_wi}) and considering $Z_0$ is known, one can
use AIS to sample $Z$ from the set of importance weights $\{\tilde \omega_i \}$.
In practice, one uses Gibbs sampling to implement the transition probabilities
$T_k({\bf x}_k,{\bf x}_{k+1})$ at each $k$, so that ${\bf x}_{k+1}$ is obtained from
${\bf x}_k$ efficiently. In this way,
one starts from a certain ${\bf x}_1$ sampled from $p_0({\bf x})$ 
and get ${\bf x}_2$
from $T_1({\bf x}_1, {\bf x}_2)$, use this new value to obtain ${\bf
  x}_3$ from $T_2({\bf x}_2, {\bf x}_3)$, and so on.
Notice once again that sampling ${\bf x}_1$ is trivial since $p_0({\bf x})$ contains only 
visible bias terms ${\bf B}$. 

In general, AIS depends on three parameters which define the
computational cost and accuracy of the approximation. A remarkable one
is the set of bias ${\bf B}$, as a suitable selection can improve the 
quality of the samples obtained, and the overall calculation.
Another relevant parameter is the 
number $N_\beta$ of intermediate probability distributions, which has to be
fixed beforehand. Finally, the number of samples $N_s$ has also to be set. 

Getting a suitable ${\bf B}$ may not be a trivial task, and in this work
we elaborate on this point. In fact, the problem of getting ${\bf B}$ can
be mapped into the problem of determining the mean value of the visible
units. This relation can be established minimizing the Kullback-Leibler (KL) 
divergence between $p_0({\bf x})$ and the full RBM probability distribution 
$p_n({\bf x})$
\[
\nabla_{\vec B} \sum_j p_n({\bf x}_j) \log\left( p_n({\bf x}_j) \over p_0({\bf x}_j) 
\right) = 0 \ ,
\]
where the sum over $j$ extends to all the $2^{Nv}$ states obtained after marginalization 
over the hidden units. With $p_0({\bf x})=2^{N_h} e^{-{\bf B}\cdot {\bf x}}/Z_0$, 
the above condition leads to
\begin{eqnarray}
0 & = & -\sum_j p_n({\bf x}_j) \nabla_{\bf B} \ln p_0({\bf x}_j)
\nonumber \\
& = & \sum_j p_n({\bf x}_j) {\bf x}_j + \sum_j p_n({\bf x}_j) \nabla_{\bf B} \log Z_0
\nonumber \\
& = & \langle {\bf x} \rangle_n - 
\left\langle {1 \over e^{\bf B} + 1} \right\rangle_n\ ,
\end{eqnarray}
where the subscript $n$ indicates that the average values are taken over the $p_n({\bf x})$ 
probability distribution corresponding to the target RBM. In this way, the optimal bias ${\bf B}$ 
are given by the expression
\begin{equation}
B_i = \log\left( {1\over \langle x_i \rangle_n} - 1 \right) 
\label{Bi_xin}
\end{equation}
for each visible unit $i\in1,2,\ldots, N_v$. The problem of finding the optimal ${\bf B}$ is
thus equivalent to obtaining the exact average values of the visible units. Since
this problem is as hard as finding $Z$ itself, one has to devise alternative strategies
to approximate $\langle x_i\rangle_n$.  

Two common strategies are usually employed to face this problem. The simplest one is to 
simply set ${\bf B}=0$ and sample from the uniform probability distribution. This is a 
very cheap procedure that has nevertheless some drawbacks as will be shown later. Another
common strategy was devised in~\cite{salakhutdinov-et-al-Annealed-Importance-Sampling-2008-ICML}, 
where the dataset used to train the RBM is employed to find $\langle {\bf x}\rangle_n$.
This last scheme, which usually works well in machine learning problems, has two potential
issues: on one hand, it can not be implemented when there is no training set; and on the other,
it assumes that the training set represents a significant subset of the highest-probability states, 
something that presumably happens only at the end of the learning process, but definitely not 
at the first epochs. The first issue is specially relevant because the problem of finding 
$Z$ is more general than its application to machine learning problems. 

In this work we introduce alternative strategies to evaluate $\langle {\bf x}\rangle_n$ that
prove to be as efficient as the previous ones, avoiding some of the commented drawbacks. In all
cases, the main idea is to estimate $\langle {\bf x} \rangle_n$ by direct sampling of the RBM.
In order to do that, two 
decisions have to be made, regarding 
the sampling scheme and the starting 
point ${\bf x}_{\rm ini}$.
As far as the sampling method is concerned, we use both Gibbs sampling and 
Metropolis sampling, the latter with an additional parameter which is the number of units
to flip at each iteration. Regarding the initial state, the number of possibilities 
is larger. We have tried a few of them: 
\begin{itemize}
    \item[a)] ${\bf x}_{\rm ini}=0$.
    \item[b)] ${\bf x}_{\rm ini}=1$.
    \item[c)] ${\bf x}_{\rm ini}$ sampled from the Bernoulli distribution ($p=0.5$).
    \item[d)] Mean field (MF) approximation for a high-probability state: in this scheme one assumes that the fluctuations of the hidden
    units with respect to their mean values is so small that one can neglect quadratic terms. One starts from the
    free energy of a given visible state and replaces ${\bf h}$ by $\langle {\bf h}\rangle + \delta{\bf h}$, expanding the 
    exponential of the $\delta{\bf h}$ terms to first order
    \begin{eqnarray}
      {\mathcal F}(\tilde{\bf x}) & = & \sum_{\tilde{\bf h}} e^{\tilde{\bf x}^{\rm T} \tilde{\bf W}\tilde{\bf h}} =
      \sum_{\tilde{\bf h}} e^{\tilde{\bf x}^{\rm T} \tilde{\bf W} (\langle \tilde{\bf h}\rangle + \delta \tilde{\bf h})} 
      \nonumber \\
      & \approx & \sum_{\tilde{\bf h}} e^{\tilde{\bf x}^{\rm T} \tilde{\bf W} \langle \tilde{\bf h}\rangle}
      \left( 1 + \tilde{\bf x}\tilde{\bf W}\delta\tilde{\bf h} \right)
      \nonumber \\ 
      & = & 2^{N_h} e^{\tilde{\bf x}^{\rm T} \tilde{\bf W} \langle \tilde{\bf h}\rangle} + 
      \tilde{\bf x} \tilde{\bf W} e^{\tilde{\bf x}^{\rm T} \tilde{\bf W} \langle \tilde{\bf h}\rangle} 
      \sum_{\tilde{\bf h}} \left( \tilde{\bf h} - \langle \tilde{\bf h} \rangle \right)
      \nonumber \\
      & = & 2^{N_h} e^{\tilde{\bf x}^{\rm T} \tilde{\bf W} \langle \tilde{\bf h}\rangle} \ .
    \end{eqnarray}
    The resulting expression does not depend on the explicit values
    of ${\bf h}$, but only on their mean values, which are always positive when working with $\{0,1\}$ units. In this way, 
    $\mathcal F({\bf x})$ is maximal when $x_i \sum_j \omega_{ij}$ is maximal $\forall i$. A necessary condition for this to happen is
    that $x_i$ must have the same sign as $\sum_j \omega_{ij}$. Since $x_i$ can not be negative, we simply set
    $x_i$ to $1$ if $\sum_j\omega_{ij}>0$, and 0 otherwise.
    
    \item[e)] Pseudoinverse (PS) approximation for a high probability state: 
    one can also look for a state of the complete (visible and hidden) space that has 
    a large probability as the starting point of the sampling process. Things are much simpler than
    in d) because in this case one works directly with the energy rather than the free energy.
    Setting to zero the gradients with respect to ${\bf x}$ and ${\bf h}$ of the energy in Eq.~(\ref{energy-RBM}), 
    one finds
    \begin{equation}
      {\bf x}_p = -({\bf W}^+)^T {\bf c} 
      \,\,\, , \,\,\,
      {\bf h}_p = -{\bf W}^+ {\bf b} \ ,
    \end{equation}
    where ${\bf W}^+$ is the pseudoinverse of the ${\bf W}$ matrix. Notice that these equations are decoupled,
    and that therefore one can start from either of them in a Gibbs sampling scheme. In this work we 
    build ${\bf x}_{\rm ini}$ from ${\bf x}_p$, by rounding it to the $\{0,1\}$ range.

\end{itemize}

In this way, the total computational cost associated to the evaluation of $Z$ is 
divided in two parts, 
one corresponding to building 
${\bf B}$,
and another,  
proportional to $N_\beta\cdot N_s$, corresponding the AIS 
algorithm. The approximation obtained improves when both $N_\beta$
and $N_s$ tend to infinity~\cite{neal-Annealed-Importance-Sampling-2001-SC}. Starting the AIS algorithm from the 
uniform probability distribution may require extremely high values of $N_\beta$ and $N_s$ to obtain
a reasonably good approximation of $Z$.
Typical values of $N_\beta$ and $N_s$
commonly employed in previous works range from a few hundreds to
several thousands, but at least one of them is always
large~\cite{salakhutdinov-et-al-Annealed-Importance-Sampling-2008-ICML,cote-larochelle-InfiniteRBM-2016-NC}.
However, to the best of our knowledge, no systematic
analysis on the convergence properties of the AIS algorithm applied to
RBMs has been carried out so far, other than the discussion in Ref.~\cite{krause-et-al-2020}.

In this work we argue that, by getting 
a good starting probability distribution $p_0({\bf x})$ from a suitable bias ${\bf B}$, 
one can obtain the same accuracy with much lower values of $N_\beta$ and $N_s$. As a
consequence, the overall computational cost, including the 
cost associated to building the ${\bf B}$'s, turns out to be much lower.
In summary, we study the compound problem of finding a good $p_0({\bf x})$ 
and evaluating AIS, to find the best possible approximation of $Z$ with
as low computational cost as possible. The $p_0({\bf x})$ probability distributions
explored are always trivially sampled as they only contain visible bias terms, which
are obtained using both Metropolis and Gibbs sampling methods as explained above.

We explore both artificial and
large-sized realistic problems. For the artificial tests, we build
systems where the exact value of the partition function can be
computed and compared with the AIS results, thus being a benchmark for
the latter. We analyze the convergence of
the AIS estimate for fixed and small $N_\beta$ and $N_s$ starting from different
$p_0({\bf x})$ distributions and sampling methods, and compare to the exact result.
From this analysis we extract a set of strategies and parameters that we use to
obtain the estimation of $Z$ in realistic problems. These results are compared with
what we take as the ground truth, which is an AIS estimate of $Z$ using
$N_\beta=2^{20}$ and $N_s=1024$, starting from the corresponding training set
and using the Salakhutdinov procedure outlined 
in~\cite{salakhutdinov-et-al-Annealed-Importance-Sampling-2008-ICML}.


A final relevant remark is to realize that the value of $Z$ is invariant under the
exchange of the visible and hidden units, provided $\tilde {\bf W}$ is transposed. 
This is not trivial from the point of view of the AIS algorithm 
since in AIS one evaluates the probabilities of the visible states, which are 
different in each case. As a consequence, AIS does not necessarily gives the same
predictions. This can be of particular relevance when the system is highly unbalanced,
as usually happens in machine learning problems, where the number of hidden units 
is typically different from the number of visible units.
For that reason we explore all the previous parameters and strategies both for 
the original and the transposed system, in the latter case including also the 
analysis for ${\bf B}=0$, corresponding to the uniform probability distribution.
Remarkably, though, the computational cost involved in AIS is the same independently of whether
$\tilde{\bf W}$ is transposed or not when Gibbs sampling is employed in the intermediate chains
of the algorithm.


As a side note, the reader may realize that in both steps (building ${\bf B}$
and performing AIS) there is at least one dimension that can be easily 
parallelized. At this level, using NVIDIA\textsuperscript{\textregistered}
CUDA\textsuperscript{\textregistered} hardware is a perfect fit for the task.

\section{Description of the Experimental Problems analyzed}


In order to perform a systematic study of the convergence properties
of the AIS algorithm, we test its performance in several problems
where one can analytically compute the exact value of the
partition function. We separate the sets of weights in two 
categories, one corresponding to purely artificial sets, and the other
to simplified realistic problems where a dimension is largely reduced.


Regarding the purely artificial systems, we explore the following cases:

\begin{itemize}

\item[1)] Gaussian Weights with Gaussian Moments (GWGM)
The first artificial model employed is characterized by a weights matrix 
of Gaussian random numbers with $N_v=20$ and two values of $N_h$, namely 60 and 180. 
The mean value and the standard deviation of each weights matrix are also 
sampled from a Gaussian distribution, as explained in the Sec.~\ref{section-experiments}.
Due to the reduced value of $N_v$, the exact calculation of $Z$ can be performed by brute force. 

\item[2)] Block Matrix Systems (BMS) \\
  In this model the weights matrix ${\bf W}$ is organized by blocks
  located at its diagonal. All blocks can be different, but they are
  of a reduced size that allows for an independent, exact evaluation
  of its corresponding partition function. In this way, the weights matrix
  takes the form
  \begin{equation}
    {\bf W} = 
    \left(
    \begin{array}{cccc}
      {\bf W}_1 & 0 & \cdots & 0 \\
      0 & {\bf W}_2 & \cdots & 0 \\
      \vdots & \vdots & \ddots & \vdots \\
      0 & 0 & \cdots & {\bf W}_{N_B}
    \end{array}
    \right) \ .
    \label{W_blocks}
  \end{equation}
  If ${\bf W}$ is formed by
  $M_B$ blocks, the partition function of the overall system is simply
  \[
  Z = \prod_{j=1}^{M_B} Z_j \ ,
  \label{Z_boxed_a1}
  \]
  with $Z_j$ is the partition function of each isolated
  block. In this work we take each ${\bf W}_k$ to be of the GWGM form.

\end{itemize}

In much the same way, the simplified realistic problems analyzed are:

\begin{itemize}

\item[3)] A set of weights obtained after training a RBM with the MNIST dataset\footnote{http://yann.lecun.com/exdb/mnist},
with $N_h=20$ hidden units (MNIST-20h). We monitor and store the weights at each epoch along the 
learning process, and make a suitable final selection that characterizes the evolution of the
weights matrix at all stages during the learning process, as described in the Sec.~\ref{section-experiments}.
In this way, we have snapshots of the system taken at the beginning of the learning, where the training
set typically does not contain the 
highest probability states, and at the end, where they are supposed to carry most of the 
probability density mass of the system. 

\item[4)] A set of weights obtained after training a RBM with the Web-w6a dataset, with $N_h=20$ hidden units 
(Web-w6a-20h) in the same conditions as the MNIST-20h case. This dataset contains data from web pages with 
$N_v=300$ sparse binary keyword attributes taken from~\cite{platt-1999-bkch}. We use this problem to train a 
RBM containing 49749 examples in the training 
set obtained joining all the available original examples. 

\end{itemize}

Finally, we test the performance of AIS with the studied strategies on realistically large-sized problems.
In particular, we analyze the MNIST and Web-w6a systems, trained with a much larger number of hidden units,
$N_h=500$ (MNIST-500h and Web-w6a-500h).

%
%
%
%
%

\section{Experimental Results}
\label{section-experiments}

\subsection{Results for Exactly Solvable Models with ${\bf B}=0$}
\label{sec_exact_solv_res_B0}

In the following we report our results regarding the AIS estimation of
the partition function for the different systems described
in the previous section, where the exact $Z$ can be numerically computed.
In all cases, the dimensions of the systems have been chosen 
such that one dimension is large and comparable to what one finds in
real-world problems.
One of the major goals of this work is to obtain good approximations of $Z$ with 
small values of the AIS parameters. Unless otherwise stated, 
the number of AIS chains and AIS samples have been set to 
$N_\beta=N_s=1024$, as an overall balance between computational cost and statistical accuracy.
These values are small compared to what one usually finds in the literature~\cite{salakhutdinov-et-al-Annealed-Importance-Sampling-2008-ICML,cote-larochelle-InfiniteRBM-2016-NC}. 
In any case, all values are taken to be a power of $2$, as we have checked 
that convergence towards the exact value seems to be logarithmic. 

We start by analyzing the ${\bf B}=0$ case, so that $p_0({\bf x})$ is 
the uniform probability distribution. This is the simplest scheme as there is no
computation of a starting bias ${\bf B}$. 
In the GWGM and BMS cases, all weights have been randomly chosen, following a Gaussian distribution.
The mean and standard 
deviation of each Gaussian has also been chosen to be Gaussian, with $\mu_\mu$ and $\sigma_\mu$ the
mean and standard deviation of the mean values, and $\mu_\sigma$ and $\sigma_\sigma$ the mean and
standard deviation of the standard deviations. In order to mimic what one finds in realistic 
learning problems, the mean and standard deviation of the bias terms can be different from the ones
corresponding to the ${\bf W}$ weights. In this work we introduce a factor $\lambda$ that scales the $\mu$ and $\sigma$ 
of ${\bf b}$ and ${\bf c}$ with respect to those of ${\bf W}$,
\[
\mu_{\rm bias} = \lambda\,\mu_{\bf W} 
\,\,\, , \,\,\,
\sigma_{\rm bias} = \lambda\,\sigma_{\bf W} \ .
\]
In the MNIST-20h and Web-w6a-20h cases the weights are the result of learning the corresponding problems
using CD$_1$, at different epochs (10, 50, 100, 200, 300, 400, 500). 

\begin{center}
\begin{figure}[!t]
\includegraphics[width=\linewidth]{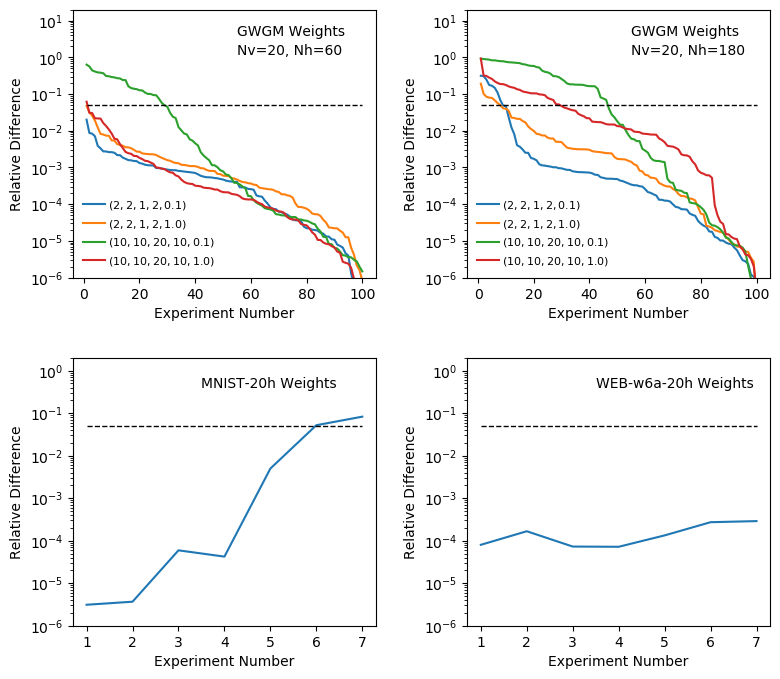}
\caption{Relative differences between the exact $\log(Z)$ and the AIS estimated $\log(Z)$ starting from 
the uniform probability distribution (${\bf B}=0$). All calculations correspond to $N_s=1024$ AIS samples
and $N_\beta=1024$ intermediate probability distributions. The horizontal dashed line indicates a relative 
difference of $5\%$, see the text for further details.}
\label{fig_AIS_p0_Uniform}
\end{figure}
\end{center}

We show in Fig.~\ref{fig_AIS_p0_Uniform} a selection of the
results obtained for some of the exactly solvable models analyzed.  
For GWGM, the specific values of $(\mu_\mu, \sigma_\mu, \mu_\sigma, \sigma_\sigma, \lambda)$ are given in the plot.
In all cases, the vertical axis indicates
the relative differences $\xi = |(\log(Z_{\rm Ex})-\log(Z_{\rm AIS}))/\log(Z_{\rm Ex})|$,
with $\log(Z_{\rm Ex})$ and 
$\log(Z_{\rm AIS})$ the exact and AIS-estimated results, respectively. For the GWGM weights, the horizontal axis
corresponds to the experiment number, sorted according to the value of $\xi$.
The MNIST-20h and Web-w6a-20h cases are sorted according to the epoch number.
In the latter cases, the estimated likelihood stabilizes in the final epochs. 
In all plots, a horizontal dashed line indicating a relative difference of $5\%$ has been included. 
We take this value as a (subjective) threshold for a suitable approximation of $\log(Z)$.

As it can be seen from the figure, two different behaviours arise. In many cases there is a large 
region where the AIS algorithm starting from the uniform probability distribution performs very well, 
leading to very small relative differences with respect to the exact values. This is particularly
notorious in the Web-w6a-20h case. In the MNIST-20h, differences arise at the end of the learning, but 
they are relatively small. On the contrary, the GWGM problems show a behaviour that 
depends on the size of the system and the parameters defining the Gaussian weights, as can be seen
in the upper panels. Notice, though, that the most divergent curves correspond to the larger systems, 
a tendency that we have verified also in the BMS case. 
One may think that the weight values in these experiments are
large (in absolute value) compared to what one usually finds in RBM learning problems. Still, 
one should keep in mind that in realistic learning problems, the tendency of the absolute value 
of the weights is to increase along learning. As an example, the Root Mean Square value of the 
MNIST-20h weights at the epochs considered is $(0.22, 0.38, 0.48, 0.59, 0.66, 0.71, 0.76)$, while for the
Web-w6a these values are smaller. This fact has to be taken into account if one seeks to perform
a calculation of $Z$ along learning, as the problem hardens as learning progresses.
Parenthetically, we have verified that, at least for the purely artificial problems, 
the agreement between $\log(Z_{\rm Ex})$ and $\log(Z_{\rm AIS})$ improves
when the parameters of the Gaussian distribution used to build the weights are very small.
The BMS systems formed by GWGM blocks show a remarkably similar behaviour to the upper panels in the
figure. Packing GWGM weights with a similar $\xi$ in a BMS structure leads to approximately the 
same value of $\xi$. In this way, building blocks out of smaller arrangements does not seem
to alter the intrinsic difficulty of the problem.

\begin{center}
\begin{figure}[!t]
\includegraphics[width=\linewidth]{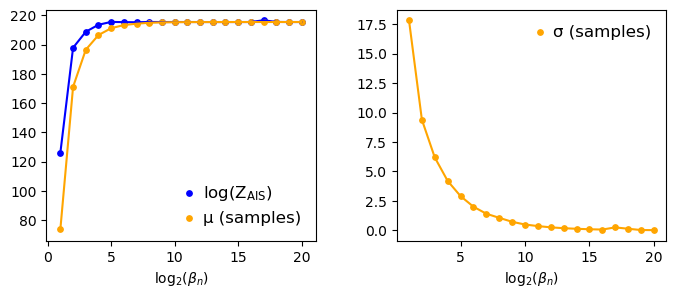}
\caption{left panel: AIS estimation of $\log(Z)$, starting from the uniform probability distribution,
for the MNIST-20h weights at epoch 500 (blue points), and mean of the samples (orange points).
Right panel: standard deviation of the same set of samples.
Results have been obtained with $N_s=1024$ samples and 
the values of $N_\beta$ span the range $(2^1,2^{20})$.}
\label{fig_MNIST-20h_AIS_p0_Uniform_Betas}
\end{figure}
\end{center}

Another potential issue of starting from the uniform probability distribution is that 
the estimation obtained 
seems to converge fast to a possibly wrong value when increasing $N_\beta$. 
Theoretical arguments indicate that, in the long run, the estimation must converge
to the exact value, but this convergence can be extremely slow. In this way, one can perform a calculation and 
get the wrong impression that the result obtained is close to the exact one.
This is shown in the left panel of Fig.~\ref{fig_MNIST-20h_AIS_p0_Uniform_Betas}, where $\log(Z_{\rm AIS})$ 
and the mean of the AIS samples for the MNIST-20h problem at epoch 500
are depicted as a function of the number of intermediate chains used, for fixed $N_s=1024$ samples.
One could infer from the plot that AIS is yielding a good and stable estimation, which is
$\log(Z_{\rm AIS}) \approx 215.17$, while the exact value is $\log(Z_{\rm Ex})=234.85$. 
To make things worse, the standard deviation of the samples decreases with increasing number of AIS 
chains and approaches zero, as shown in the right panel of the figure. 
Notice that not even with $N_\beta=2^{20}$ the AIS prediction changes appreciably to approach the 
exact value. This reinforces the false impression 
that $\log(Z)$ is accurately estimated. Although not reported here,
we have seen that this is a common behaviour found in other problems. 

In summary, the analyzed set of exactly solvable models indicates that, while the uniform probability
distribution may seem to be a good starting point for AIS, this is not the case in general.
In fact, using a finite but large number of 
AIS chains does not guarantee that one approaches the exact result. Even worse, one can get the false
impression that already a simple calculation with a small number of chains leads to a stable result
close to the exact one (as shown in Fig.~\ref{fig_MNIST-20h_AIS_p0_Uniform_Betas}), while this may
definitely not be the case.
We have checked that this situation is not solved by increasing the number of samples, either.
Depending on the desired accuracy,
these facts call for a better initialization of the AIS algorithm. We elaborate on this point in the next
sections.

\subsection{Analysis of the Transposed System for ${\bf B} = 0$ in Exactly Solvable Models}
\label{sec_exact_solv_res_B_0_against_B_0_Transposed}

\begin{center}
\begin{figure}[!t]
\includegraphics[width=\linewidth]{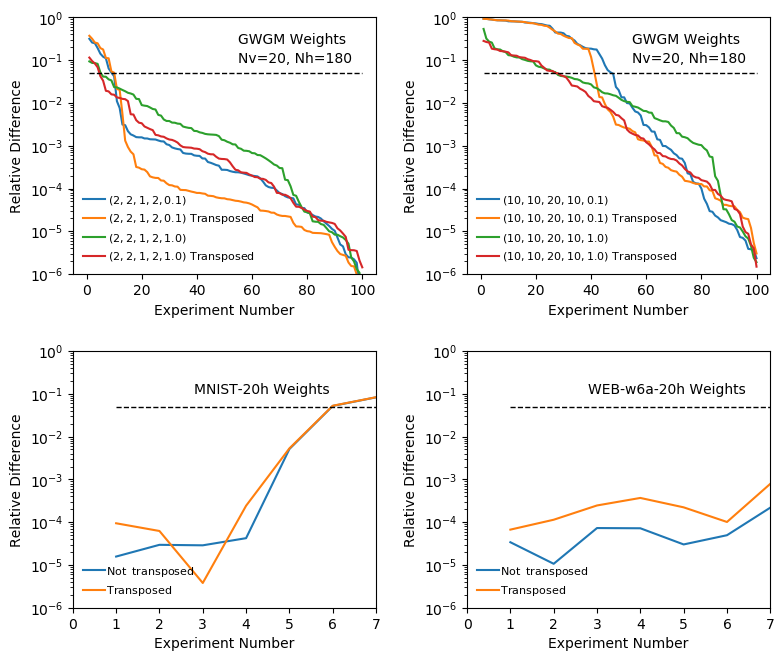}
\caption{Relative differences $\xi$ between $\log(Z_{\rm Ex})$ and $\log(Z_{\rm AIS})$ starting
from the uniform probability distribution (${\bf B}=0$) for transposed and non-transposed systems.
All calculations correspond to $N_s=1024$ AIS samples and $N_\beta=1024$ intermediate probability
distributions. The horizontal dashed line indicates a relative difference of $5\%$.}
\label{fig_p0_against_p0_Transposed}
\end{figure}
\end{center}

We start by analyzing the simplest alternative, which is to transpose the system while keeping ${\bf B=0}$.
Transposing the system implies exchanging the visible and hidden layers, and to replace $\tilde{\bf W}$
with its transpose $\tilde{\bf W}^T$. 
 For the sake of comparison, we show in Fig.~\ref{fig_p0_against_p0_Transposed} 
the performance of AIS when the original and the transposed system is sampled from the uniform
probability distribution, for the most difficult artificial cases of Fig.~\ref{fig_AIS_p0_Uniform}
in the upper panels, and for the MNIST-20h and the Web-w6a-20h in the lower panels. 
As in Fig.~\ref{fig_AIS_p0_Uniform}, a horizontal dashed line indicating the value
$\xi=0.05$ is also displayed.
Results show very little improvement since the most marked differences appear for quite
low values of $\xi$, whereas for larger $\xi$'s the differences are less significant.
Note that in the purely artificial systems, in contrast to the MNIST-20h and Web-w6a-20h cases,
the number of visible units in the transposed system is much larger than in the original one. 
Still, none of the plots indicate that transposing leads to a statistically meaningful 
better solution when starting from the uniform probability distribution. As we will show later,
this is not the case when using other strategies and parameters. In any case, transposing is 
a very simple operation that does not substantially change the AIS complexity, does not seem to
worsen the estimation, and that can always be considered as a cheap alternative to the standard 
procedure.

\subsection{Results for Exactly Solvable Models with ${\bf B}\neq 0$}
\label{sec_exact_solv_res_B_non_0}

\begin{table*}[t!]
\begin{center}
\begin{tabular}{|l|l|l|l|l|} \hline
Strategy        & \#Samples & \#Steps & Number of Bit Changes & $\epsilon$             \\ \hline
Gibbs sampling  & 1024      & 1, 10, 100      & -                     & 0.01, 0.05, 0.10, 0.20 \\ \hline
Metropolis      & 1024      & 10, 100         & $5\%, 10\%, 15\%, 25\%, 50\%, 100\%$ (GWGM and BMS) & 0.01, 0.05, 0.10, 0.20  \\ 
                &           &                 & 1, 10, 40, 80 (MNIST-20h and Web-w6a-20h)        &  \\ \hline
\end{tabular}
\caption{Description of the parameters used in the experiments for the two strategies employed.}
\label{table_parameters_experiments_B_non_0}
\end{center}
\end{table*}

In this section we explore and discuss the different AIS predictions one obtains starting from
a non-uniform probability distribution given by ${\bf B}\neq 0$. The bias ${\bf B}$ are obtained
using the strategies (Gibbs Sampling and  Metropolis) and the initializations outlined in
section~\ref{section-AIS_RBM}. 
Each strategy has its own parameters, and we vary them over a wide range of values in order to 
determine their relevance on ${\bf B}$ and on the resulting $\log({\rm Z}_{\rm AIS}$). 
Two common parameters for both strategies are the number of samples and the number of steps between samples.
In the Metropolis case there is an additional parameter, which is the number of bits to flip in each proposal. 
Eventually, these values can depend on the size of the problem, and we report the ones we have used in 
table~\ref{table_parameters_experiments_B_non_0}.
In each case we generate a set of samples that are used to 
compute the mean values $\langle x_i\rangle_n$ used in Eq.~(\ref{Bi_xin}). In practice, though, 
Eq.~(\ref{Bi_xin}) can not be directly used as $\langle x_i\rangle_n=0$ and $\langle x_i\rangle_n=1$ give
divergent bias. In this way, one has to regularize that expression, introducing a cutoff value 
to avoid this problem. In this work we use a cutoff $\epsilon$ to linearly rescale the $[0,1]$ 
interval into the $[\epsilon, 1-\epsilon]$ one, with $\epsilon$ ranging from 0.01 to 0.20, as
shown in Table~\ref{table_parameters_experiments_B_non_0}.

Armed with all these combinations, we perform a search of the best-overall sets that improve
as much as possible over the ${\bf B}=0$ case. The aim of this search is to find a combination
of values that works well in most cases, and that can be used as a suitable starting point
for any AIS calculation. Of course, one can not guarantee that these values will be the best ones
for any specific problem, but we expect them to work well in similar cases. 


\begin{center}
\begin{figure}[!t]
\includegraphics[width=\linewidth]{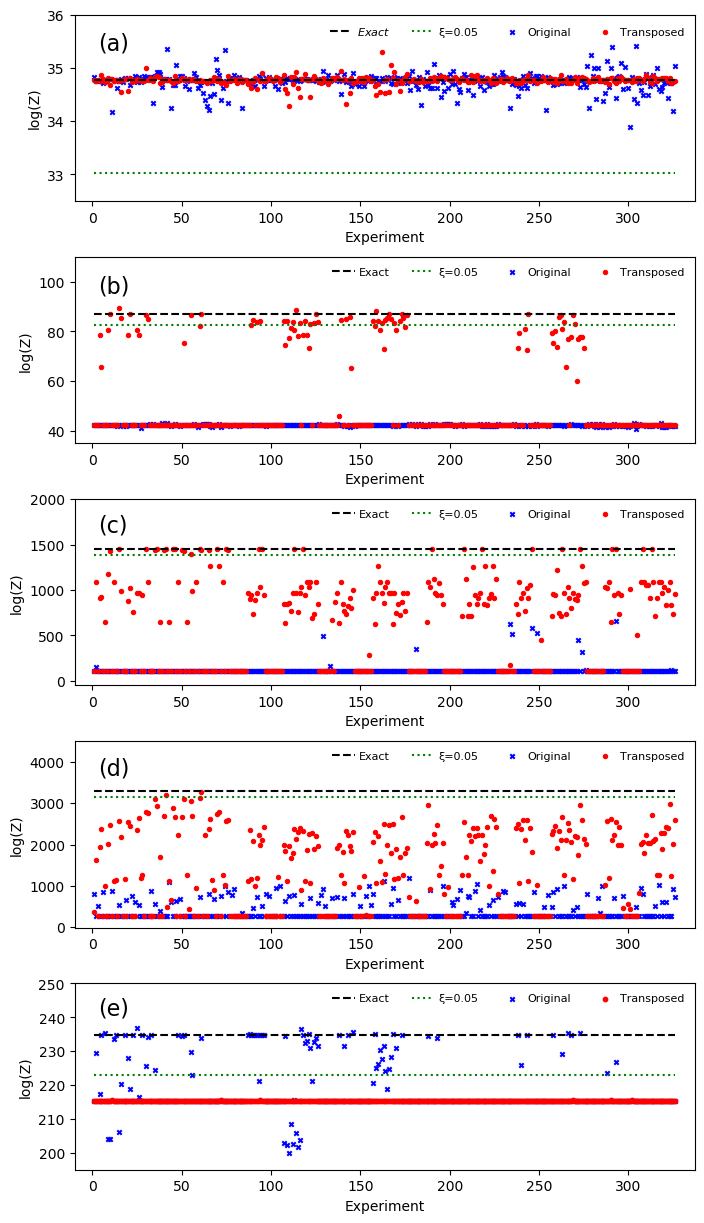}
\caption{Results of all possible strategies, initializations and parameters for the 
selected sets of weights in figure~\ref{fig_AIS_p0_Uniform} (see text for details). Black dashed line: exact
value of $\log(Z_{\rm Ex})$; green dotted line: value of $\log(Z_{\rm AIS})$ that leads to a relative
difference of $5\%$; blue crosses: original weights; red circles: transposed weights.} 
\label{fig_Lokos_STRATEGIES}
\end{figure}
\end{center}

In order to find the best possible combinations of strategies, initializations and
parameters, we perform an exhaustive analysis of all possible 325 combinations 
in several problems, including the GWGM, BMS and the MNIST-20h datasets.
For the GWGM case we have selected 
three sets of weights extracted from the most difficult curve of Fig.~\ref{fig_AIS_p0_Uniform},
corresponding to $N_v=20,\,\, N_h=180$ and $(\mu_\mu, \sigma_\mu, \mu_\sigma, \sigma_\sigma, \lambda) = (-10, 10, 20, 10, 0.1)$.
These sets yield a value of $\xi=0.92,\,0.55,\,0.06$ in the curve. 
The BMS case is built from three GWGM weights with $\xi=0.923, 0.922$ and $0.904$.
For the MNIST-20h case, we have selected the set of weights at 500 epochs, where the ${\bf B}=0$ estimations
are worse.
Figure~\ref{fig_Lokos_STRATEGIES} shows the AIS results,
obtained from all possible strategies and parameters of table~\ref{table_parameters_experiments_B_non_0} 
and initializations of section~\ref{section-AIS_RBM},
for the GWGM (panels (a), (b) and (c)), BMS (panel (d)) and MNIST-20h (panel (e)) sets of weights.
The black dashed line in the plots indicates the value of $\log(Z_{\rm Ex})$, 
while the green dotted one corresponds to $\xi=0.05$ as in Fig.~\ref{fig_AIS_p0_Uniform}.

As it can be seen, for the GWGM problem all combinations work in the easy case of panel (a), while the accuracy 
in the intermediate and hardest ones (panels (b) and (c)) is dramatically reduced. In this sense, there is a
strong correlation between difficulty and the number of working combinations.
Remarkably, transposing the weights leads to overall better estimations. In fact, in almost all cases 
the estimation for the transposed system is closer to $\log(Z_{\rm Ex})$ than the original one, a behaviour
that is not reproduced when starting from ${\bf B}=0$. In this sense and for these problems, 
transposing seems to be a proper choice when combined with the proposed strategies.
This may be related to the fact that, in the cases analyzed, transposing makes the number of visible
units to be larger than the number of hidden units, providing more degrees of freedom to both ${\bf B}$ 
and the AIS samples. The same conclusions apply to the BMS plot of panel (d), as the transposed system is 
remarkably better predicted than the original one.
The MNIST-20h points in panel (e) show that this problem is not easy, as 
with the GWGM cases in panels (b) and (c). In this case, transposing the weights
leads to the dramatically stable and bad prediction of around 215.17 of Figs.~\ref{fig_AIS_p0_Uniform}
and \ref{fig_MNIST-20h_AIS_p0_Uniform_Betas}. Again, this can be traced back to the
fact that the original system already has $N_v=784\gg N_h=20$ units.

\begin{center}
\begin{figure}[!t]
\includegraphics[width=\linewidth]{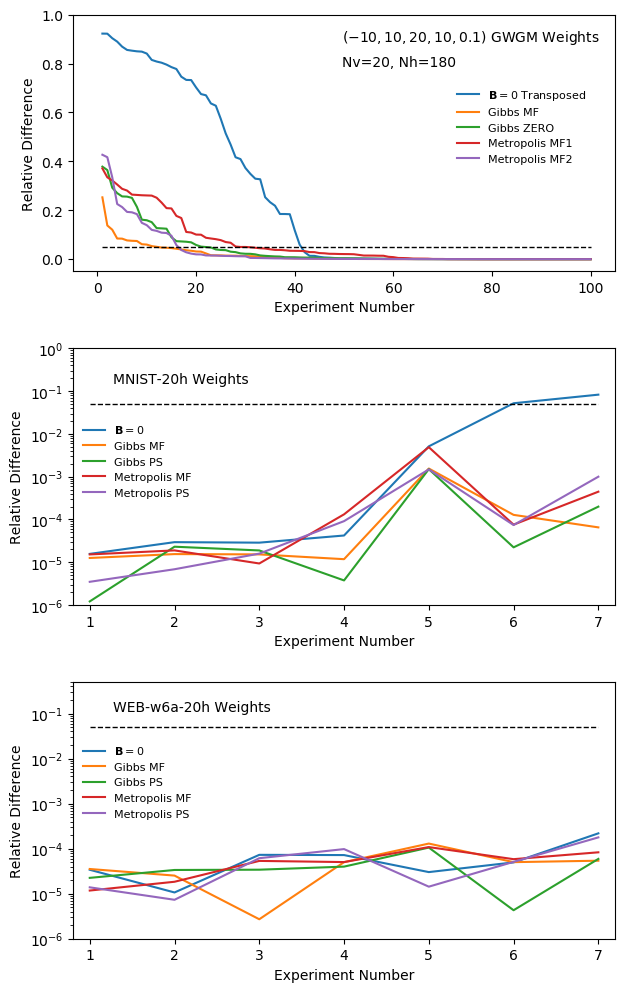}
\caption{Results of the selected good working strategies for GWGM (upper panel), MNIST-20h (middle panel)
and Web-w6a-20h (lower panel) weights, see text for details. The black dashed line indicates a relative
difference $\xi=5\%$ with respect to $\log(Z_{\rm Ex})$.}
\label{fig_Best_of_Best_Strategies}
\end{figure}
\end{center}

\begin{table*}[t!]
\begin{center}
\begin{tabular}{|l|l|l|l|l|l|} \hline
Problem     & Strategy+Initialization  & \#Samples & \#Steps & Number of Bit Changes & $\epsilon$  \\ \hline
GWGM        & Gibbs MF                 & 1024      & 1       & -                     & 0.10        \\ 
            & Gibbs ZERO               & 1024      & 1       & -                     & 0.10        \\ 
            & Metropolis MF1           & 1024      & 10      & 15\%                  & 0.05        \\ 
            & Metropolis MF2           & 1024      & 10      & 100\%                 & 0.10        \\ \hline
MNIST-20h   & Gibbs MF                 & 1024      & 100     & -                     & 0.05        \\ 
            & Gibbs PS                 & 1024      & 100     & -                     & 0.05        \\ 
            & Metropolis MF            & 1024      & 100     & 1                     & 0.10        \\ 
            & Metropolis PS            & 1024      & 100     & 1                     & 0.10        \\ \hline
Web-w6a-20h & Gibbs MF                 & 1024      & 100     & -                     & 0.05        \\ 
            & Gibbs PS                 & 1024      & 100     & -                     & 0.05        \\ 
            & Metropolis MF            & 1024      & 100     & 1                     & 0.10        \\ 
            & Metropolis PS            & 1024      & 100     & 1                     & 0.05        \\ \hline
\end{tabular}
\caption{Description of the parameters used in the best strategies found for very problem.}
\label{table_parameters_Best_of_Best}
\end{center}
\end{table*}

Despite the fact that one can find good strategies for specific sets of weights, it is very unlikely
to get  overall winning strategies that work well in all cases. In fact, the physics described by 
arbitrary weights can be very different, ranging from paramagnetic to ferromagnetic states or 
spin glasses. The selection of good strategies, therefore, is similar to the model selection procedure in
a typical Machine Learning scenario. In this way, an exhaustive search should be performed if one
seeks to obtain the best possible estimation of $\log(Z)$. We show in Fig.~\ref{fig_Best_of_Best_Strategies}
a selection of good working strategies for the 
$N_v=20,\,\, N_h=180$, $(\mu_\mu, \sigma_\mu, \mu_\sigma, \sigma_\sigma, \lambda) = (-10, 10, 20, 10, 0.1)$
GWGM weights, together with the MNIST-20h and the Web-w6a-20h ones. 
In each case we have chosen the two Gibbs sampling and two Metropolis strategies 
that work best for that particular problem, with the corresponding 
parameters reported in table~\ref{table_parameters_Best_of_Best}. As in the previous plots, we 
also include the prediction for the ${\bf B}=0$ system of Fig.~\ref{fig_p0_against_p0_Transposed}, 
and indicate with a black dashed line a relative difference limit of $5\%$ with respect to the exact value
of $\log(Z)$. 
Notice that the GWGM weights have been transposed, while the original configurations have been kept for
the MNIST-20h and Web-w6a-20h sets.
On the one hand, one readily notices that the selected strategies work very well in Web-w6a-20h
problem, as does the ${\bf B}=0$ estimation. On the other hand, the MNIST-20h is well reproduced
by the selected strategies, 
even at the final epochs where the ${\bf B=0}$ estimations fails.
The most complex scenario is found on the GWGM case, where at least a 40\% of the predictions made by 
the ${\bf B}=0$ AIS calculations lay above the accuracy threshold of $5\%$.
In this case one observes a neat improvement of the selected strategies along two different directions:
first, the number of estimations above the $5\%$ limit is reduced, and second,
the relative differences $\xi$ are also lower than in the ${\bf B}=0$ case.

In summary, for all the problems at hand, one can always find a combination of 
strategies, initializations and parameters that give a better AIS prediction that
any estimation obtained starting from ${\bf B}=0$.
In this sense, one may wonder whether it is possible to find a set of strategies that generically perform better than
the ${\bf B}=0$ ones in most cases. As shown above, the best strategies are problem-dependent, a situation
that is difficult to overcome and common to other Machine Learning situations.

From our experiments we have observed that Metropolis sampling can perform better than Gibbs sampling, at the cost
of choosing the value of an additional parameter, namely the number of units to flip at each step. Furthermore, 
the cutoff $\epsilon$ also plays a significant role, while it is meant to avoid singular values of ${\bf B}$ but 
not to introduce or modify the underlying physics of the problem. Regarding the initializations, the most successful ones 
correspond to the MF or PS choices, which is an interesting feature as they depend explicitly on the matrix of weights.
Finally, transposing seems to be the best choice when the number of hidden units in the original system is larger than the 
number of visible ones. Based on these observations and for the sake of completeness,
we have selected a couple of strategies, Gibbs$_{\rm mf}$ and Gibbs$_{\rm ps}$,
that perform better than the uniform probability distribution
in general, and that could be taken as the starting point 
in generic AIS calculations. Both strategies are based on Gibbs sampling, and use $\epsilon=0.05$ to avoid excessive
modification of the mean values $\langle x_i\rangle_n$ of Eq.~(\ref{Bi_xin}). Finally and as mentioned above, one uses the MF
initialization and the other uses PS. In both, the number of intermediate steps between samples have been set to 100, and the 
number of samples to average have been set to 1024. In much the same way, the number of AIS chains and AIS samples is always set 
to 1024 as the main goal of this work is to obtain good approximations of $\log(Z)$ with low computational cost.

\begin{center}
\begin{figure}[!t]
\includegraphics[width=\linewidth]{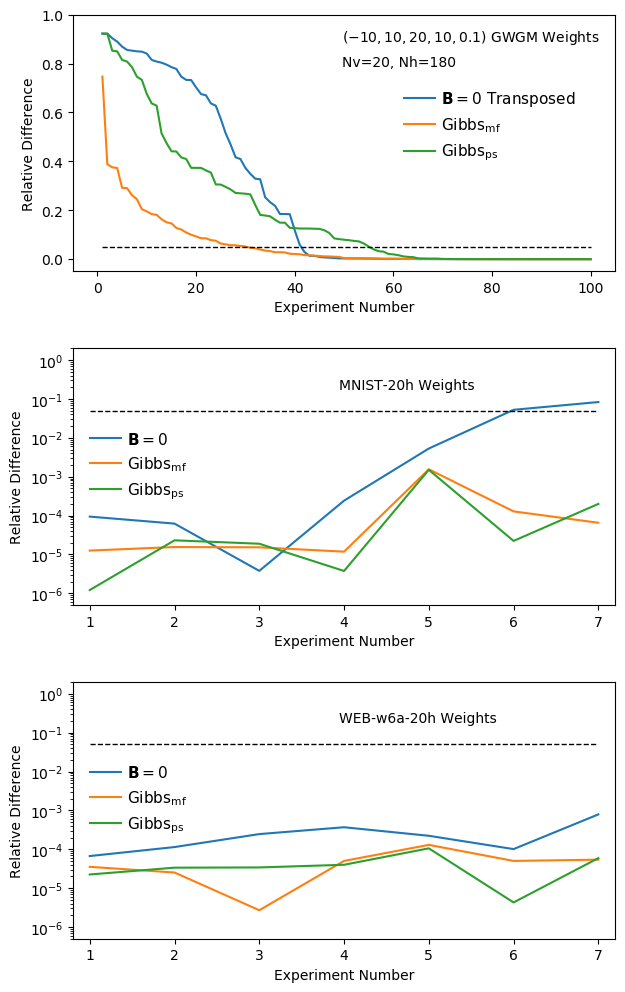}
\caption{Results of the two selected strategies Gibbs$_{\rm mf}$ and Gibbs$_{\rm ps}$ for GWGM (upper panel),
MNIST-20h (middle panel) and Web-w6a-20h (lower panel) weights, as explained in the text. The black dashed
line indicates a relative difference $\xi=5\%$ with respect to $\log(Z_{\rm Ex})$.}
\label{fig_Selected_Strategies}
\end{figure}
\end{center}

Results from these two strategies are compared in Fig.~\ref{fig_Selected_Strategies} with the prediction of 
the ${\bf B}=0$ system for the GWGM weights with $N_v=20,\,\, N_h=180$ and
$(\mu_\mu, \sigma_\mu, \mu_\sigma, \sigma_\sigma, \lambda) = (-10, 10, 20, 10, 0.1)$, the MNIST-20h and the Web-w6a-20h cases 
(upper, middle and lower panels, respectively).
As can be seen from the figure, both strategies perform well in the MNIST-20h and Web-w6a-20h cases, and better than
the uniform probability distribution. Both strategies improve also over the ${\bf B}=0$ estimation for the GWGM case, 
but in different ways. The MF curve improves along the two directions previously mentioned, as there are few sets of weights
with a relative difference error $\xi<5\%$, while the rest have a $\xi$ value that is clearly lower than 
the one of the ${\bf B}=0$ curve. On the contrary, the PS curve presents more cases with $\xi>5\%$, although most of these
have a lower value than the ${\bf B}=0$ prediction. 
In any case, a relative difference error close to $5\%$ is still affordable in most situations, so a small increment
in the number of cases around that value is less dramatic than accepting larger errors. In this way, we understand than the 
PS curve is, overall, better than the ${\bf B}=0$ one.
According to these results we conclude that, at least for the problems analyzed in this work, the MF strategy is the
preferred one. We believe that it can be used as a suitable starting point for AIS calculations in other problems.

\begin{center}
\begin{figure}[!t]
\includegraphics[width=\linewidth]{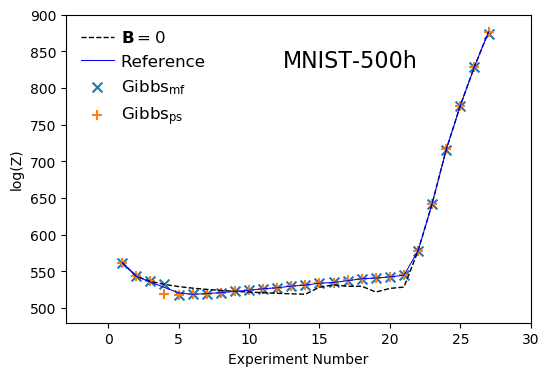}
\caption{Results of the two selected strategies Gibbs$_{\rm mf}$ and Gibbs$_{\rm ps}$ for the MNIST-500h
weights at different epochs along learning.}
\label{fig_Selected_Strategies_MNIST_500h}
\end{figure}
\end{center}

Armed with the selected strategies, we move to the more realistic Machine Learning, non-artificial MNIST-500h and
Web-w6a-500h problems. For these systems there is no exact calculation of $\log(Z)$ and one has to rely to the values
obtained from state-of-the-art techniques found in the literature. For that matter we take as reference the value obtained from 
the procedure of Ref.~\cite{salakhutdinov-et-al-Annealed-Importance-Sampling-2008-ICML}, which uses the dataset 
employed to train the RBM as the starting point to compute the mean values required in the evaluation of ${\bf B}$ 
of Eq.~(\ref{Bi_xin}). With this ${\bf B}$, we run AIS with $N_s=1024$ and $N_{\beta} = 2^{20}$ to obtain the reference value.

We show in Fig.~\ref{fig_Selected_Strategies_MNIST_500h} the reference values, the ${\bf B}=0$ estimation and the prediction
of our two selected strategies for the MNIST-500h problem and a few selected epochs along the learning process.
The first twenty points correspond to the first twenty epochs where $\log(Z)$ rapidly evolves, while the last points
correspond to epoch numbers $40, 100, 200, 300, 400$ and $500$. As it can be seen, all curves merge at the highest epochs, 
while the ${\bf B}=0$ prediction departs from the reference curve at the early and intermediate epochs. On the contrary, the 
selected strategies are hardly distinguishable from the reference line along the whole curve. The same comparison for the Web-w6a-500h
shows excellent agreement between all curves, including the ${\bf B}=0$ prediction.

\section{Summary and Conclusions}

In summary, in this work we analyze the performance of the AIS algorithm in the evaluation of the partition function $Z$ of a 
Restricted Boltzmann Machine with a reduced number of samples and intermediate chains. We evaluate $\log(Z)$ for 
a number of exactly solvable models, including Gaussian weights, Block Matrix Systems and realistic problems
(MNIST and Web-w6a) with a reduced number of hidden units. In particular we show that a suitable starting probability distribution
$p_0({\bf x})$ can lead to a big improvement of the AIS estimation of $\log(Z)$ for fixed number of samples and intermediate chains. 
We build $p_0({\bf x})$ from a RBM with bias terms only, and show that these values are directly related to the averages of
the visible states. We note that the reference distribution of Ref.~\cite{salakhutdinov-et-al-Annealed-Importance-Sampling-2008-ICML}
is directly related to our procedure when the training set is employed to evaluate the required averages. 
Remarkably, our methodology can be used when no training set is available.

For that matter, we propose sample the RBM, where only the weights and bias are required. 
Two standard sampling techniques are tested,
Metropolis and Gibbs sampling, with different initialization schemes and cutoff parameters. We show that suitable
combinations of parameters can be found such that the AIS estimation of $\log(Z)$ improves over the one obtained
from the uniform probability distribution. However, while the best combinations appear to be problem dependent,
we find that there are a few of them that work well in all the problems tested. We select two strategies based on
Gibbs sampling that represent a trade-off between simplicity, reduced computational cost, and accuracy. We
finally test them on the MNIST and Web-w6a with 500 hidden units to show that the estimations obtained are in
excellent agreement with the ones from the procedure in Ref.~\cite{salakhutdinov-et-al-Annealed-Importance-Sampling-2008-ICML}.
These predictions are different from the results one gets starting from the uniform probability distribution.
We expect that the strategies proposed can be used as the starting point in further studies of
$\log(Z)$ in RBMs with the AIS algorithm. 

Our analysis shows that the accuracy of the AIS prediction can vary considerably depending on the set of
weights employed. In much the same way, the distribution of the samples obtained in AIS shows different
structures for different set of weights. In fact, we have seen that the system may undergo one or more thermal phase
transitions along the AIS simulation, as the $\beta_k$ parameters can be understood as the inverse of the system's
temperature at each step. For this reason and as a future line of work, we believe that it could be interesting to
adapt the initialization algorithm to the different physical scenarios such that one starts from the proper physical
phase described by the set of weights.

\section*{Acknowledgments}
FM: This work has been supported by MINECO grant
FIS2017-84114-C2-1-P from DGI (Spain).
ER: This work was partially supported by MINECO project
PID2019-104551RB-I00 (Spain). Part of the hardware used for this
research was donated by the NVIDIA\textsuperscript{\textregistered} Corporation.

\bibliographystyle{IEEETrans/IEEEtran}
\bibliography{bibliography-paper}

\end{document}